\documentclass{article} 
\usepackage{iclr2018_conference,times}
\usepackage{times}
\usepackage{hyperref}
\usepackage{url}

 \usepackage{amsmath}
 \usepackage{bm} 
 \usepackage{graphicx}
 \usepackage{comment}
 \usepackage{makecell}
 
 \DeclareMathOperator*{\argmax}{arg\,max}

\title{Multi-modal Geolocation Estimation using Deep Neural Networks}

\author{Jesse M. Johns, Jeremiah Rounds  \& Michael J. Henry \\
Pacific Northwest National Laboratory\\
Richland, WA 99352, USA \\
\texttt{\{jesse.johns,jeremiah.rounds,michael.j.henry\}@pnnl.gov} \\
}

\begin{document}

\maketitle

\begin{abstract}
Estimating the location where an image was taken based solely on the contents of the image is a challenging task, even for humans, as properly labeling an image in such a fashion relies heavily on contextual information, and is not as simple as identifying  a single object in the image. Thus any methods which attempt to do so must somehow account for these complexities, and no single model to date is completely capable of addressing all challenges. This work contributes to the state of research in image geolocation inferencing by introducing a novel global meshing strategy, outlining a variety of training procedures to overcome the considerable data limitations when training these models, and demonstrating how incorporating additional information can be used to improve the overall performance of a geolocation inference model. In this work, it is shown that Delaunay triangles are an effective type of mesh for geolocation in relatively low volume scenarios when compared to results from state of the art models which use quad trees and an order of magnitude more training data.  In addition, the time of posting, learned user albuming, and other meta data are easily incorporated to improve geolocation by up to 11\% for country-level (750 km) locality accuracy to 3\% for city-level (25 km) localities.
\end{abstract}

\section{Introduction}

Advancements in deep learning over the past several years have pushed the bounds of what is possible with machine learning beyond typical image classification or object localization. Whereas prior work in computer vision and machine learning has focused primarily on determining the contents of the image--``is there a dog in this photo?"---more recent methods and more complicated models have enabled deeper examinations of the contextual information behind an image. This deeper exploration allows a researcher to ask more challenging questions of its models. One such question is, ``where in the world was this picture taken?"

However, estimating the geographic origin of a ground-level image is a challenging task due to a number of different factors. Firstly, the volume of available images with geographic information associated with them is not evenly distributed across the globe. This uneven distribution increases the complexity of the model chosen and complicates the design of the model itself. Furthermore, there are additional challenges associated with geo-tagged image data, such as the potential for conflicting data (eg. incorrect geolabels and replica landmarks - St Peter’s Basilica in Nikko, Japan) and the ambiguity of geographic terms (eg. imagery of similar and ambiguous geological features, such as beaches, screenshots of websites, or generally, plates of food).

The work presented herein focuses on the task of content-based image geolocation: the process of identify the geographic origin of an image taken at ground level. Given the significant growth in the production of data, and trends in social data to move away from pure text-based platforms to more images, videos, and mixed-media, tackling the question of inferring geographic context behind an image becomes a significant and relevant problem. Whereas social textual data may be enriched with geolocation information, images may or may not contain such information; EXIF data is often stripped from images or may be present but incomplete. Within the context of mixed-media data, using the geolocation information associated with a linked object may be unreliable or not even applicable.  For example, a user on Twitter may have geolocation information enabled for their account, but may post a link to an image taken from a location different from the one where they made their post.

\subsection{Related Work}
A wide variety of approaches have been considered for geolocation from image content, please consult \citep{Brejcha2017} for a review. The approach taken in this paper builds on recent work in global geolocation from ground-based imagery \citep{weyand2016planet} for global scale geo-inferencing. \cite{weyand2016planet} utilize a multi-class approach with a one-hot encoding. A common approach, as well, is to perform instance-level scene retrieval in order to perform geolocation of imagery \citep{hays2008im2gps,hays_large-scale_2015,vo_revisiting_2017}.  These works query on previously geotagged imagery and assign a geolabel based on the similarity of the image query to the database.  Further, \cite{vo_revisiting_2017} builds on work by \cite{hays2008im2gps,hays_large-scale_2015,weyand2016planet} by utilizing the feature maps for the mesh-based classifier for the features of their nearest-neighbors scene retrieval approach.

Prior work exists for data sampling strategies for large-scale classification problems for social media applications. \cite{kordopatis2016depth} considers weighted sampling of minority classes. In this work, the class selection is biased during training of the deep learning models.  Biasing class selection with random noising (often called image regularization) is a well known way to allow a model to see more examples of rare classes, but also there are additional concerns related specifically to social media applications.  For example, researchers in \cite{kordopatis2016depth} consider the case of sampling such that an individual user is only seen once per a training epoch.  In this work, sampling is performed without respect to user, so that images are selected in epochs completely at random, but the broader influence of latent variables, other than user, and communities is of concern in social media geolocation.

\subsection{Contributions}

The first type of model considered in this work is purely geolocation for image content (M1).  The use of time information forms a second set of models within this work (M2). User-album inputs form a third set of models within this work (M3).   Our work contributes meaningful consideration of the use of an alternative mesh for geolocation in M1 and demonstrates the meaningful use of time and user information to improve geolocation (M2 and M3).

\section{Geolabeled Data}
\label{geo_data}

Collected data holdings are derived from YFCC100M \citep{thomee_yfcc100m:_2016}, where training and validation was performed on a randomly selected 14.9M (12.2M training/2.7M validation) of the 48.4M geolabeled imagery.  PlaNet, in comparison, used 125M (91M training/34M validation) images \citep{weyand2016planet} for model developemt.  In using the data, it is assumed that the ground truth GPS location is exactly accurate, which is a reasonable approximation based on the results of \cite{hauff_study_2013}.

\begin{figure}
\centering
\includegraphics[scale=.11]{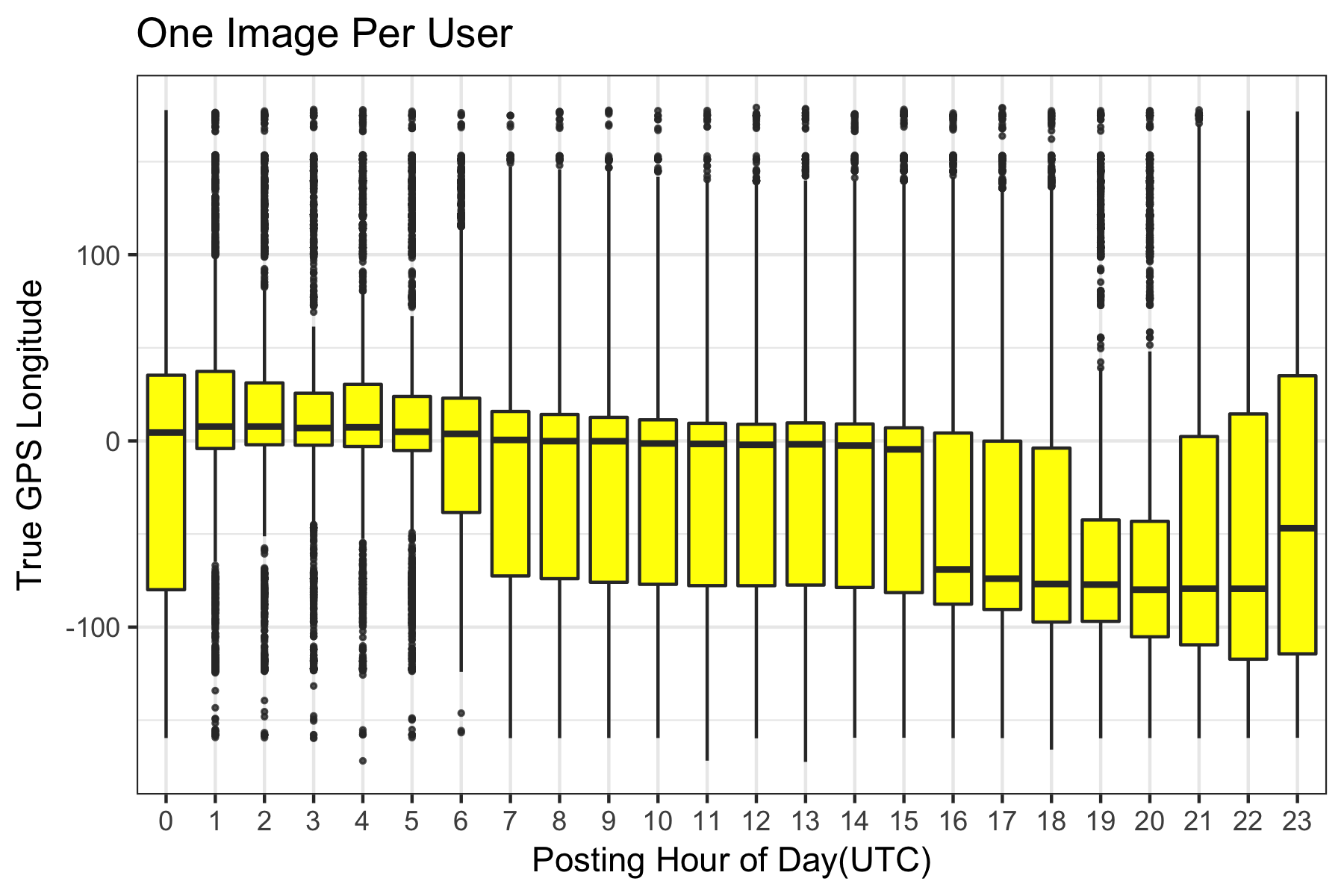}
\caption{True Longitude versus Time of Posting.  In these boxplots, we limit users to one image selected at random to decouple any user effects from time of day. What this image implies is that the prior distribution for image longitude differs by the time of day that an image is posted.   As an example, consider longitudes near 0.  It would take less evidence to predict a longitude close to zero for an image posted around 01:00 UTC than to make a similar prediction at 21:00 UTC because of the observed priors.  At 21:00 UTC, the boxplot indicates images are typically as far from zero longitude as will been seen in the data.  }
\label{fig:one-image-lng}
\end{figure}

Every YFCC100M image has associated meta data.  Importantly, these data contain user-id and posted-time. User-id is a unique token that groups images by account.   Posted time is the time that the user uploaded data to the website that ultimately created the combined storage of all images.   The most likely true  GPS  location for an image from varies by the time the image is uploaded to a website, as shown in Figure \ref{fig:one-image-lng}.

\section{Methodology}
\label{approach}

\subsection{CNN Classification (Model M1)}

The approach described in this work for the global-scale geolocation model is similar to the classification approach taken by PlaNet \citep{weyand2016planet}. Spatially the globe is subdivied into a grid, forming a mesh, decribed in Section \ref{subsec:mesh}, of classification regions.  An image classifier is to trained to recognize the imagery whose ground truth GPS is contained in a cell and hence, during inference, the mesh ascribes a probability distribution of geo-labels for the input imagery.

\subsubsection{Mesh}
\label{subsec:mesh}

The classification structure is generated using a Delaunay triangle-based meshing architecture.  This differs from the PlaNet approach, which utilizes a quad-tree mesh.  Similarly to PlaNet, the mesh cells are generated such that they conserve surface area, but the approach is not as simple as with a quad-tree.  The triangular mesh was deployed under the hypothesis that the Delaunay triangles would be more adaptive to the geometric features of the Earth. Since triangular meshes are unstructures, they can more easily capture water/land interfaces without the additional refinement needed by quad-tree meshes.  However, the triangular mesh loses refinement level information which comes with a structured quad-tree approach, which would allow for granularity to be controlled more simplistically, based which cells contain other cells.  In order to control mesh refinment, cells are adapatively refined, or divided, when the cell contains more than some number of examples (refinement limit) or the mesh cell is dropped from being classified by the model (the contained imagery is also dropped from training) if the cell contains less than a number of samples (minimum examples).  The options used for generating the three meshes in this paper are shown in Table \ref{mesh_opts}.  Parameters for the initialization of the mesh were not explored; each mesh was initialized with a 31 x 31  structured grid with equal surface area in each triangle.

\begin{figure}[h!]
\centering
\includegraphics[scale=.05]{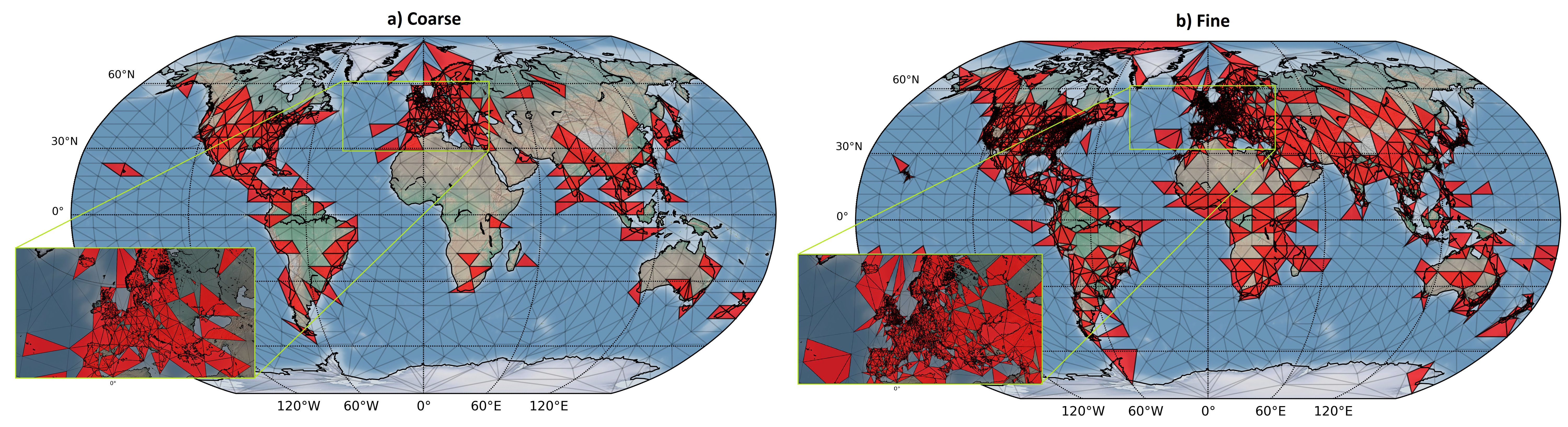}
\label{fig:mesh}
\caption{Coarse mesh and fine mesh classification structure for classification models developed in this paper. The red triangles indicate regions where the number of images per cell meets the criteria for training.  The coarse mesh was generated with an early dataset of 2M YFCC100M images, but final training was performed by populating those cells with all training data without additional refinement. The fine and fine$_P$ meshes were generated with all available imagery (14M).  The fine mesh does a better job at representing geographic regions, for example, the water/land interface of Spain and Portugal becomes evident.  However, this is masked with the coarser mesh, which lacks fidelity to natural geographic boundaries.}
\end{figure}

The mesh for the a) coarse mesh and b) fine mesh are shown in Figure \ref{fig:mesh}.

\begin{table}[!h]
\centering
\caption{The geolocation classification mesh can be tuned by modifying the refinement level to adjust the maximum and minimum number of cells in each mesh cell. This table shows the meshing parameters used for the three meshes studied in this work.  These meshes were selected to cover a reasonable range of mesh structures, with fine$_P$ meant to replicate the mesh parameters utilized by PlaNet.  However, it should be noted that this isn't direcly comparable since the fine$_P$ mesh is generated with both a different methodology and a different dataset.}
\label{mesh_opts}
\begin{tabular}{rlllr}
\hline
& \thead{Mesh}   & \thead{Refinement Limit}   & \thead{Minimum Examples} & \thead{Cell Count}  \\
\hline
&Coarse & 8000 & 1000 & 538 \\
&Fine & 5000  & 500 & 6565      \\
&Fine$_P$   & 10000 & 50 & 4771
\end{tabular}
\end{table}

\subsubsection{Model}

The Inception v4 convolutional neural network architecture proposed in \citep{szegedy_inception-v4_2016} is deployed to develop the mesh-based classification geolocation model (Model M1) presented in this work, which differs from the Inception v3 architecture used in PlaNet \citep{weyand2016planet}.  A softmax classification is utilized, similar to the PlaNet approach. One further difference between the work presented here and that of PlaNet is that cells are labeled based the cell centroid, in latitude-longitude. Here an alternate approach is used where the ``center-of-mass" of the training data in a containing cell is computed, whereby the geolocation for each cell was specified as the lat/lon centroid of the image population. Significant improvements are expected for coarse meshes, as it can be noticed in Figure \ref{fig:mesh} for example, that the cell centroid on the coast of Spain is out in the ocean.  Therefore, any beach image, as an example, will have an intrinsically higher error that would be otherwise be captured by a finer mesh.  This is expecially true for high density population regions.

\subsubsection{Model Evaluation}

Models are evaluated by calculating the distance between the predicted cell and the ground-truth GPS coordinate using a great-circle distance:

$ d = R \times arcos\left( sin(x_1) sin(x_2) + cos(x_1) cos(x_2) cos(y_2 - y_1) \right)$

where $ x_i $ and $ y_i $ is the latitude and longitude, respectively, in radians, and $ R $ is the radius of the Earth (assumed to be 6372.795 km). An error threshold here is defined as the number of images which are classified within a distance of $d$.  Error thresholds of 1 km, 25 km, 200 km, 750 km, 2500 km are utilized to represent street, city, region, country, and continent localities to remain consistent with previous work \citep{hays2008im2gps,hays_large-scale_2015,weyand2016planet,vo_revisiting_2017}.  As an example, if 1 out of 10 images are geolabeled with a distance of less than 200 km, then the 200 km threshold would be 10\%.

\subsection{Geolocation Improvement with Time Meta Data (Model M2)}

Every YFCC100M image has associated meta data; importantly these data consist of user id and posted time. Posting time is utilized in model M2, Figure \ref{tab-model-describe}. Let $Z_{ik}$ be the one-hot encoding of the i$^{th}$ image for the k$^{th}$ class, so that it takes on value 1 if image $i$ belongs to the k$^{th}$ geolocation class. $P(Z_{ik} = 1| X_{i} )$ is assumed not equal to  $ P(Z_{ik} = 1 | X_{i}, t_i)$, where $t_i$ is the time of posting of the ith image.    Evidence for this fact is of the form of observed true GPS longitude shown as changing distribution with respect to posting time (Hour of Day UTC) as shown in Figure \ref{fig:one-image-lng}.  Figure \ref{fig:one-image-lng} is only strong evidence for $P(Z_{ik} = 1| t_{i}) \neq P(Z_{ik} = 1)$, and it could still be the case that  $P(Z_{ij} = 1| X_{i} )=  P(Z_{ij} = 1 | X_{j}, t_j)$.  Which is to say, conditioned on the content of an image there could be no dependence on time, but it seems prudent with the evidence in this figure to proceed under the assumption that there is time dependence.  The operational research hypothesis for this model (M2) is that there remains a time-dependence after conditioning on image content.

To incorporate time, related variables are appended to the output of the geolocation model (M1) to form a new input for M2.  Every image has a vector of fit probabilities $\bm {\hat p_i}$ from the softmax layer of M1.  $\bm {\hat p_i}$ is filtered so that only the top 10 maximum entries remain, and the other entries are set to 0 ($\bm {\hat p^0_i}$).  This vector is normalized, so that , $\hat p^1_{ik} = \hat p^0_{ik}/\sum_l \hat p^0_{il}$ . The L1-norm (denominator) in the normalization $|\bm \hat p^0_{i}|$ is appended as a feature.  Time of posting is turned into a tuple of numeric input as $\bm T_i= ( \text{Hour}(t_i) - 11.5)/24,  (\text{Day of Week}(t_i) - 3.5)/7, (\text{Month}(t_i) - 6.5)/12$, where $t_i$ is the time of publishing, $ \text{Hour}(t_i)$ is a function that returns hour of the day (0 to 23), $\text{Day of Week}$ is a function that returns 1 to 7 for numbered days of the week starting with Sunday as 1, and $\text{Month}(t_i)$ is a function that returns the month of year starting with January as 1.  Geolocation uses as input for each image, the concatenated vector $(\bm \hat p^1_{i}, |\bm \hat p^0_{i}|, \bm T_i)$.  Year is specifically omitted from time inputs because this would not generalize to new data posted in the future.

\begin{table}
\caption{M2: Time-Adjusting Geolocation Model.  N is the length of  $\bm {\hat p_j}$, which takes on two values: 538 (coarse mesh) and 6565 (fine mesh). All layers except the last are dense layers with 50\% drop-out and batch normalization.}
\label{tab-model-time-adjust}
\centering
\begin{tabular}{llll}
\hline Layer & Type & Number Neurons & Activation \\
\hline1&Input & N+4 & ReLU\\
2&Hidden& 542*3 & Sigmoid\\
3&Hidden& 542*3 & ReLU\\
4&Hidden& 542*3& Sigmoid \\
5&Output& N& Softmax  \\\hline
\end{tabular}
\end{table}

\subsection{User-Album Geolocation Refinement (Model M3)}

\begin{figure}
\centering
\includegraphics[scale=.3]{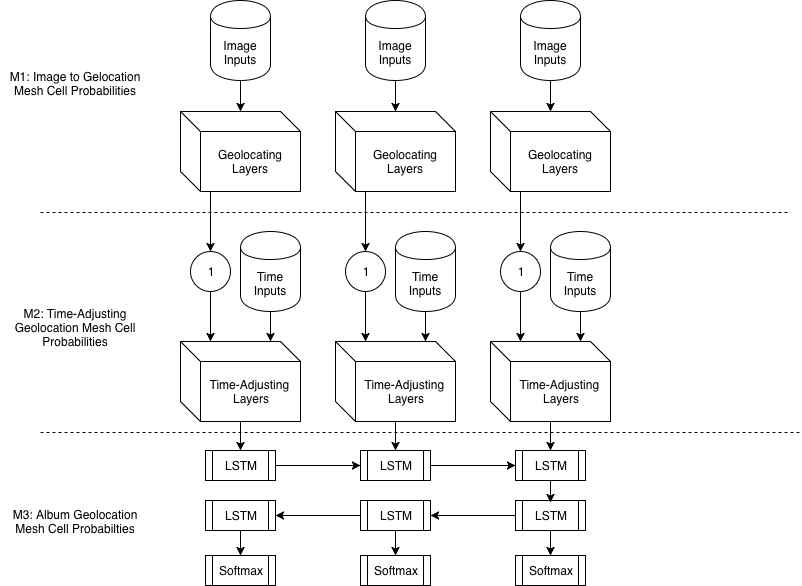}
\caption{Album model with component details added.  Operator 1 is a filtering and re-normalization of M1: Geolocation output.  At operator (1) the output of M1 is filtered to output only the top 10 mesh cell probabilities (making it mostly sparse) re-normalized to sum to 1.  Training of M2 and M3 was only done on the validation data of M1 (using a new random split of test-validation).  Time inputs are concatenated to filtered and normalized outputs at (1) and a new training step is implied.  A small abuse of notation is present: M3 (Time-Albums) is properly described  as M2 concatenated with M3 layers.  }
\label{tab-model-describe}
\end{figure}

Model M3, Figure \ref{tab-model-describe}, simultaneously geolocates many images from a single user with an LSTM model.   The idea here is to borrow information from other images a user has posted to help aid geolocation.The Bidirectional LSTMs capitalizes on correlations in a single user's images. LSTMs were also considered by \cite{weyand2016planet}, but in PlaNet the albums were created and organized by humans.  When a human organizes images they may do so by topic or location.  In M3, all images by a single user are organized sequentially in time with no particular further organization.   The related research question is: does the success observed by \cite{weyand2016planet}  extend to this  less informative organization of images?   All images from a user are grouped into albums of size 24.  If there are not enough images to fill an album then albums are 0 padded and masking is utilized.  During training, a user was limited to a single random album per epoch.

Album averaging was considered by \cite{weyand2016planet} (PlaNet).  In album averaging, the images in an album are assigned a mesh cell based on the highest average probability across all images. This method increases accuracy by borrowing information across related images.   As a control, a similar idea is applied to user images, in which, the location of an image is determined as the maximum average probability across all images associated with the posting user.  This result assumes that all images from a user are from the same mesh grid cells. In addition with user-averaging, there is no optimization that controls class frequencies to be unbiased.

Finally,  LSTM on a time-ordered sequence of images was considered (without respect to user).  However, we were unable to improve performance significantly past that gained by just adding time to the model, so albums without user are not further considered in this paper.

\section{Results}

\subsection{Model M1 Classification Experiments}

\subsubsection{Mesh Comparison}

Meshing parameters are investigated to understand the sensitivity to mesh adaptation.  The results for each mesh is shown in Table \ref{mesh_table}.  There is an apparent trade-off between fine-grain and coarse-grain geolocation.  The coarse mesh demonstrates improved large granularity geolocation and the fine mesh performs better at finer-granularities, as would be expected.  This observation was also noticed in \citep{vo_revisiting_2017}.  In addition, the impact of classifying on the centroid of the training data is compared to utilizing the cell centroid for class labels.  A dramatic improvement is noticed for the coarse mesh, with only a modest improvement for the fine mesh.

\begin{table}[h!]
\centering
\caption{Geolocation locality error thresholds on the YFCC100M validation hold-out set comparing both imagery centroid and cell centroid classification structures.}
\label{mesh_table}
\begin{tabular}{rllllllr}
\hline
& \thead{Mesh} &  \thead{Imagery Centroid} &\thead{Street \\1 km} & \thead{City\\25 km} &
 \thead{Region\\200 km}  & \thead{Country\\750 km} & \thead{Continent\\2500 km} \\
\hline
&Coarse  & False          & 0.42 & 6.89  & 21.22  & 38.53  & 56.64   \\
&Coarse   & True          & 1.25 & \textbf{10.57}  & \textbf{23.99}  & \textbf{38.72} & \textbf{56.66} \\
&Fine    & False          & 3.20 & 9.56  & 16.64 & 29.49 & 49.17  \\
&Fine     & True          & \textbf{4.44} & 10.49 & 16.63 & 29.57 & 49.13 \\
&Fine$_P$  & False          & 1.86 & 8.23 & 16.23 & 30.98 & 50.97 \\
&Fine$_P$  & True           & 3.23 & 9.47  & 16.69 & 31.07 & 50.98
\end{tabular}
\end{table}
\subsubsection{Outdoor Imagery}

An experiment was conducted to determine if the performance of the M1 model could be improved by applying the geolocation model on outdoor imagery only.  From \cite{thomee_yfcc100m:_2016} it would be expected that approximately 50\% of the geolabeled YFCC100M imagery contains outdoor imagery.  A Place365 \citep{zhou_places_2016} model was used in conjunction with indoor-outdoor label deliminations to filter geolocation inference only on outdoor imagery.  Note that the geolocation model was not re-trained on the ``outdoor" imagery, this is only conducted as a filtering operation during inference. Results are shown in Table \ref{outdoor_table}.  In general, the improvement is quite good, about a 4-8\% improvement in accuracy for region/country localities, with a more modest boost in smaller-scale regions.

\begin{table}[h!]
\centering
\caption{Geolocation locality error thresholds on the YFCC100M validation hold-out set comparing inference on outdoor imagery only.}
\label{outdoor_table}
\begin{tabular}{rllllllr}
\hline
& \thead{Mesh} &  \thead{Outdoor Only} &\thead{Street \\1 km} & \thead{City\\25 km} &
 \thead{Region\\200 km}  & \thead{Country\\750 km} & \thead{Continent\\2500 km} \\
\hline
&Coarse & False           & 1.25 & 10.57  & 23.99  & 38.72 & 56.66 \\
&Coarse & True            & 1.46 & 12.15 & \textbf{29.22} & \textbf{46.23} & \textbf{63.50} \\
&Fine   & False           & 4.44 & 10.49 & 16.63 & 29.57 & 49.13 \\
&Fine   & True            & \textbf{4.90} & \textbf{12.17} & 19.96 & 34.84 & 54.33 \\
&Fine$_P$ & False           & 3.23 & 9.47  & 16.69 & 31.07 & 50.98 \\
&Fine$_P$ & True            & 3.57 & 10.82 & 19.85 & 36.2 & 55.94
\end{tabular}
\end{table}

\subsubsection{IM2GPS Test Results}

The Im2GPS testing data is utilized to test the model on the 237 images provided by the work of \cite{hays2008im2gps}.  Results are tabluated in Table \ref{img2gps_results} for all of the meshes.  The imagery centroid classification labels are generated with the YFCC100M training data, yet the performance is still greatly improved when applied to the Im2GPS testing set, demostrating generality with the approach.  The performance of the M1 classification model is comparable to the performance of \citep{weyand2016planet} with a factor of ten less training data and far fewer classification regions (6k compared to 21k); the coarse mesh M1 model exceeds the performance of PlaNet for large regions (750 and 2500 km).

\begin{table}[h!]
\caption{Geolocation locality error thresholds on the Im2GPS testing data comparing both imagery centroid and cell centroid classification structures. Bold indicates the best of the models developed in this work. The work of \cite{weyand2016planet} used 125M images for model development (91M training/34M validation) and \cite{vo_revisiting_2017} is an instance retrieval based method, utilizing a PlaNet-inspired mesh-based classifier for feature extraction which was trained on 9M Flickr images.}
\centering
\label{img2gps_results}
\begin{tabular}{rllllllr}
\hline
& \thead{Mesh} &  \thead{Imagery Centroid} &\thead{Street \\1 km} & \thead{City\\25 km} &
 \thead{Region\\200 km}  & \thead{Country\\750 km} & \thead{Continent\\2500 km} \\
\hline
&Coarse  & FALSE & 0.71 & 11.35 & 29.32 & 55.62 & \textbf{72.02} \\
&Fine    & FALSE & 4.57 & 20.41 & 32.78 & 49.95 & 67.06 \\
&Fine$_P$  & FALSE & 3.27 & 16.79 & 28.87 & 46.01 & 64.04 \\
&Coarse  & TRUE  & 2.89 & 18.47 & \textbf{35.53} & \textbf{56.17} & \textbf{71.99} \\
&Fine    & TRUE  & \textbf{7.53} & \textbf{22.42} & 33.82 & 50.96 & 67.07 \\
&Fine$_P$  & TRUE  & 4.75 & 18.73 & 30.20 & 46.23 & 64.10 \\
&\cite{hays2008im2gps}& & & 12.0 & 15.0 & 23.0 & 47.0 \\
&\cite{hays_large-scale_2015}& & 2.5 & 21.9  & 32.1  & 35.4  & 51.9 \\
&\cite{weyand2016planet}& & 8.4 & 24.5 & 37.6 & 53.6  & 71.3 \\
&\cite{vo_revisiting_2017}& & 12.2 & 33.3 & 44.3 & 57.4  & 71.3
\end{tabular}
\end{table}

\subsection{Time Adjusted Geolocation and Album Results (M2 and M3)}

Use of time improves geolocation in two ways.  There is a slight gain of accuracy  for using time as indicated in Tables \ref{outdoor_table} and \ref{tab:time-acc}.  24.20\% of images are geolocated to within 200 km as opposed to 23.99\% without using time with the coarse mesh, and  12.28\% versus 10.49\% are within 25 km using the fine mesh.  This small persistent advantage can be seen across all error calculations and is statistically significant.

There is a measurable difference between the error of the coarse mesh using time (M2) and not using time (M1). There exist a matched pair of errors for each image: $e^1_i$ and $e^2_i$, where $i$ is a validation image index.  The first superscript being the M1 error and the second superscript being M2 error for image $i$ in km.  A null hypothesis is $\mu_1 = \mu_2$ where $\mu_1$ is the mean of the unknown distribution of $ e^1_i$ and likewise for $\mu_2$.  This hypothesis is tested with a Wilcoxon-Signed-Test for paired observations, which makes a minimal number of distributional assumptions. Specifically, normality of the errors is not assumed. The difference $\hat \mu_2 - \hat \mu_1 = 381$ is highly significant (p-value  $< 10^{-16}$ ) in favor of using time-inputs, so even though the effect of M2 is small, it is not explained by chance alone.

It is the case that the distribution of errors is mean shifted, but it is not uniformly shifted to lower error, nor is it it the case that images are universally predicted closer to the truth.     The median of coarse mesh $e^1$ is 1627 km while the median of $e^2$ is 1262 km  (Table \ref{tab:bias}).

Time input models appear to have lower-bias class probabilities.  Cross-entropy was optimized in training for both the classification model (M1) and time-inputs models (M2).  In each training method the goal is to minimize these class biases.     $\mbox{KL-Divergence}(\hat p, \hat q)$ is calculated for each model, where p is the observed class proportions for the true labels in validation, and q is the observed class proportions for the model predicted labels in validation (in both cases, 1 is added to the counts prior to calculating proportions as an adjustment for  0 count  issues). The KL-divergence of the model output class frequencies compared to the true class frequencies in validation are in Table \ref{tab:bias}.

``User-Averaging" is incorporated into results because it is a simple method  that appears to be more accurate then predicting individual images with M1 or M2; however, it biases cell count frequency (Table \ref{tab:bias}). In general when using the average probability vector to predict a user's image, there is no guarantee that the class frequencies are distributed similarly to the truth; thus, improved accuracy can come with higher bias which is what is observed. Albums are a much better approach to borrow information across users because built into the training method is a bias reducing cross-entropy optimization, and indeed LSTMs on user albums had the lowest class bias of any model considered.

\begin{table}[ht]
\small
\caption{Percentage accuracy to specific resolution: Researched models compared at various spatial resolutions.  Coarse and fine mesh have 538 and 6565 triangles across the globe (cells), respectively.  "Time inputs" indicates time meta information has been concatenated with M1 output.  Albums are created using user-id and contain 24 images.  Bold is the best observed accuracy in column. Coarse Mesh and Fine Mesh Best Possible are not actual models, but rather the best possible accuracy if every image was given exactly the right class.  In the fine mesh the best possible accuracy is incidental, but in the coarse mesh it is a severe limitation for street level accuracy.}
\label{tab:time-acc}
\centering
\begin{tabular}{rllllllr}
  \hline
 & \thead{Model} &  \thead{Outdoor Only} &\thead{Street \\1 km} & \thead{City\\25 km} &
 	\thead{Region\\200 km}  & \thead{Country\\750 km} & \thead{Continent\\2500 km} \\ \hline
   & Coarse Mesh Time Inputs (M2) & False &1.20 & 10.47 & 24.20 & 40.12 & 60.66 \\
    & Coarse Mesh Time Inputs (M2)  & True &  1.36 & 11.99 & 29.28 & 47.60 & 66.69 \\
    & Coarse Mesh User Averaging  & False & 0.84 & 11.34 & 27.30 & 44.74 & 62.84 \\
   & Coarse Mesh User Albums (M3) & False & 1.25 & 11.25 & 25.65 & 42.38 & 63.17\\
      & Coarse Mesh User Albums (M3)&True & 1.39 & 12.52 & \textbf{30.24} & \textbf{49.17} & \textbf{68.61} \\
     & Coarse Mesh Best Possible (cell centroid) & False & 1.57 & 25.33 & 81.97 & 99.99 & 100.00 \\
     & Coarse Mesh Best Possible (imagery centroid) & False & 4.10 & 43.38 & 93.86 & 99.98 & 100.00 \\
   & Fine Mesh Time Inputs (M2) & False & 4.82 & 12.28 & 18.44 & 31.39 & 52.25 \\
      & Fine Mesh Time Inputs (M2) & True  & \textbf{4.95} & \textbf{13.44} & 21.17 & 35.85 & 56.31 \\
  & Fine Mesh Best Possible (cell centroid) & False & 21.92 & 73.13 & 97.26 & 99.99 & 100.00 \\
  & Fine Mesh Best Possible (imagery centroid) &  False & 31.22 & 83.23 & 99.09 & 99.99 & 100.00 \\
   \hline
\end{tabular}
\end{table}

\begin{table}[ht]
\caption{KL-Divergence of various models from observed class proportions in validation data.  Lower KL-Divergence is better, and 0 indicates distributional equivalence between model class frequencies and true class frequencies.  Bold is best model observed by these measures.  Entropy has no "best", so it is not bolded.}
\label{tab:bias}
\centering
\begin{tabular}{lrrrr}
  \hline
 \thead{Model} & \thead{KL\\Divergence} & \thead{Shannon\\Entropy} & \thead{Mean\\Average Error (km)} & \thead{Median\\Image Error (km) } \\
  \hline
   Coarse Mesh Inception (M1)   & 0.6718 & 5.6360 & 3897 & 1627 \\
    Coarse Mesh Time Inputs (M2) & 0.4998 & 5.2890  & 3516 & 1262 \\
    Coarse Mesh User Averaging  & 0.9702 & 5.0760 & 3192 & \textbf{1051} \\
  Coarse Mesh User Albums Time Inputs (M3)  & 0.4991 & 5.1900 & 3320 & 1124 \\
  Coarse Mesh Best Possible & 0.0000 & 6.0180 & 62 & 33 \\
  Fine Mesh Inception (M1) & 1.7120 & 7.0460 & 4561 & 2705 \\
  Fine Mesh Time Inputs (M2) & \textbf{0.3543} & 8.2770 & 4311 & 2140 \\
  Fine Mesh Best Possible &  0.0000 & 8.6090 & 17 & 4 \\
   \hline
\end{tabular}
\end{table}

\section{Conclusions}

Conditioning on latent variables can only improve geolocation models.  Universally using time of day in the models was observed to increase accuracy and lower bias.  Time of day is a weak addition to Inception-like results, but it is useful to be as accurate as possible, and it makes statistically significant improvements to geolocation.  Both meshes were improved by using time information (M2).   This is a result that is not surprising, and as a research approach, can be applied to any number of meta data variables that might accompany images.

Accounting for indoor/outdoor scenes in images explained variation in validation accuracy. Outdoor only results are better than results for all images.  We suggest as future work that the probability that an image is outdoor could be concatenated to the input of M2.  The accompanying research hypothesis is that in the case that an image is indoors, perhaps the model will learn to weight time or other meta data results more heavily, or otherwise be able to use that information optimally.

Increasing the granularity of a grid reduces accuracy at the country and regional level, while improving accuracy at street and city level accuracy.  To be clear though, street level geoinferencing is not practical with a course mesh.  This is shown by the best possible accuracy in Table \ref{tab:time-acc}, and so it is expected that a fine mesh would do better.  On the other hand, there is no reason to assume that a fine mesh has to be better for geolocation at larger resolutions than 25 km, nor is there any explicit way to prove a fine mesh should do no worse than a course mesh. What we observe is that a course mesh is a superior grid for 200 km resolutions. Furthermore, we show that for both the coarse mesh and the fine mesh, using a Delaunay triangle-based mesh provides the ability to train accurate models with far fewer training examples than what was previously published.

\section*{Acknowledgments}

This research was performed at Pacific Northwest National Laboratory, a multi-program national laboratory operated by Battelle for the U.S. Department of Energy.

\bibliography{iclr2018_conference}
\bibliographystyle{iclr2018_conference}

\newpage
\appendix

\section{Training Procedure}

Images were divided at random into training and validation sets of 12.2M and 2.7M images and associated metadata, respectively.  Validation data used for M1 was further sub-divided at random into training and validations sets for training time-based models (M2 and M3), so that no data used to train M1 was used to also train M2 and M3.

\subsection{Model M1}
Historically, softmax classification is shown to perform quite poorly when the number of output classes is large \citep{bengio_quick_2003,mnih_scalable_2009,shen_self-organized_2017}. During initial experiments with large meshes ($ > $ 4,000 mesh regions), a training procedure was developed to circumvent these challenges.   Empirically, this procedure worked for training the large models presented in this paper; however, it is not demostrated to be ideal or that all steps are necessary for model training.  This approach started by training the model, pre-trained on ImageNet \citep{russakovsky_imagenet_2014}, with Adagrad \citep{duchi_adaptive_2011}. Second, the number of training examples were increased each epoch by 6\%, with the initial number of examples equal to the number of cells times $ 200 $. Third, the classes were biased by oversampling minority classes, such that all classes were initially equally represented.  A consequence of this approach, however, is that the minority classes are seen repeatedly and therefore the majority classes have significantly more learned diversity.  Fourth, the class bias was reduced after each model had completed a training cycle --- previous weights loaded and the model re-trained with a reduced bias. The final model was trained with SGD, using a linearly decreasing learning rate reduced at 4\% with each epoch, without class biasing and with the full dataset per epoch.  The initial value of the learning rate varied for each model (between 0.004 and 0.02). The values of those hyperparameters were empirically determined.

\subsection{Model M2 and M3}
The layers of M2 are described in  Table \ref{tab-model-time-adjust}. M2 is trained using He intializations (\citep{HeZR015}), initial iterations of Adaboost\citep{freund1995desicion}, followed by ADAM at learning rates of 0.005 and 0.001 \citep{kingma2014adam}.  Early stopping is used to detect  a sustained decrease in validation accuracy \citep{caruana2001overfitting}.

\section{M1 Feature-based Image Retrieval}

\begin{figure}[h!]
\centering
\includegraphics[scale=.2]{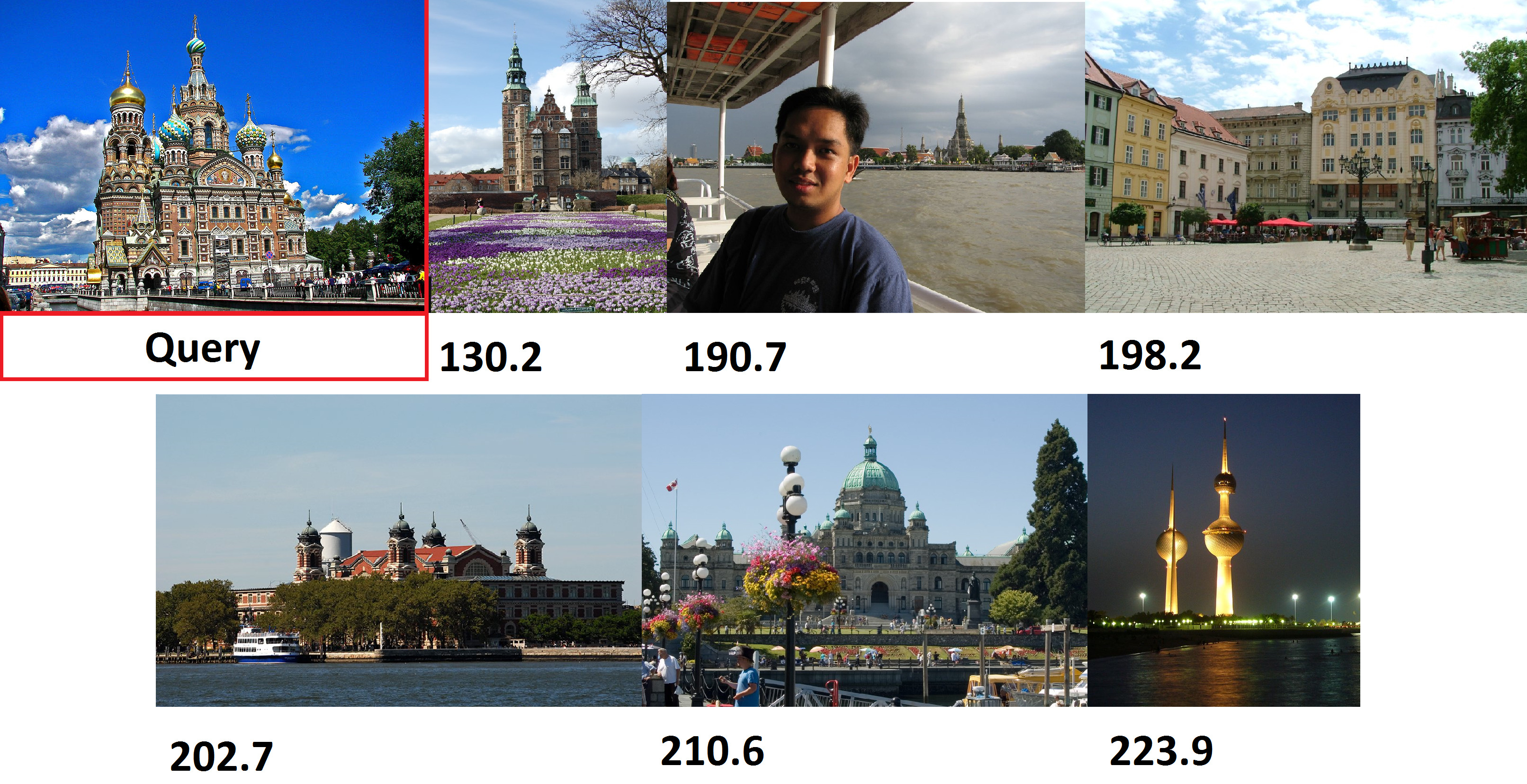}
\caption{Fine-mesh based query-by-example on Im2GPS 2K random testing dataset for a qualitative assement for model generalization. The L2 distance is computed and shown below the top six queried results.}
\label{fig:query_example}
\end{figure}

The generality of the M1 classification model is demonstrated by performing a query-by-example on the 2K random Im2GPS.  An example of an image of the Church of the Savior on Spilled Blood is shown in Figure \ref{fig:query_example}.  By manual inspection (querying on the bounding box by GPS location), this church was not present in the training data nor in the 2K random dataset \citep{hays2008im2gps}.

\section{Class Bias Analysis}

Each image is given categorical indicator variable $z_{ik} = 1$ if the $i^{th}$ image is in the kth class, 0 otherwise.   There exist a latent class distribution $ p_i = P(Z_i = 1 )$ which is assumed constant between training, testing, and application. An estimate of this unknown distribution is $ \hat p_z =\dfrac{1}{N} \sum_{j \in \text{Training}} I(z_{ij} = 1)$.  The second to last layer in all trained networks is assumed to be a fit logit for class $i$: $L_i = (L_{1i}, ..., L_{Ki})$ for the $i^{th}$ image where $K$ is the number of classes in the mesh grid.  The last layer output from networks is a softmax $\hat p^m_{{ik}} = exp(L_{ik})/\sum_v {exp(L_{iv})}$, where the $m^{th}$ index indicates model output given appropriate image input.  Optimization is done by minimizing the cross entropy between $\hat p_i$ and $\hat p^m_{i}$  as $-\sum_k \hat p_k log \hat p^m_{{ik}}$.   Images are given the class $\hat Y_{i} = \argmax_v    \hat p^m_{v}$, $\hat z_{ik} = I(\hat Y_i = k)$.

The observed distribution of class frequencies in validation is $\hat q_k =\dfrac{1}{|\text{Validation}|} \sum_{k \in \text{Validation}} I(\hat z_{ik} = 1)$. As a diagnostic we investigated how close  the class frequencies are to each other when both are calculated from the validation data. In general if two models are compared we prefer the most accurate, but may also tilt toward unbiased models in classification distribution.  If training has been done well, it should be the case that the KL-divergence between $q$ and $p$ is low: $\sum_k \hat p_k log \dfrac{\hat p_k}{\hat q_k}$. As a matter of completeness we also consider the entropy of $p$ and $q$.

\end{document}